\pdfoutput=1

\documentclass[11pt]{article}

\usepackage[preprint]{acl}

\usepackage{times}
\usepackage{latexsym}

\usepackage[T1]{fontenc}

\usepackage[utf8]{inputenc}

\usepackage{microtype}

\usepackage{inconsolata}

\usepackage{graphicx}

\title{Probing Context Localization of Polysemous Words in Pre-trained Language Model Sub-Layers}

 \author{Soniya Vijayakumar, Josef van Genabith \and Simon Ostermann \\
         German Research Instutite for Artificial Intelligence (DFKI), \\ Saarland Informatics Campus, Germany \\
soniya.vijayakumar, josef.van\_genabith, simon.ostermann  @dfki.de }

\begin{document}
\maketitle
\begin{abstract}
In the era of high performing Large Language Models, researchers have widely acknowledged that contextual word representations are one of the key drivers in achieving top performances in downstream tasks.
In this work, we investigate the degree of contextualization encoded in the fine-grained sub-layer representations of a Pre-trained Language Model (PLM) by empirical experiments using linear probes. Unlike previous work, we are particularly interested in identifying the strength of contextualization across PLM sub-layer representations (i.e. Self-Attention, Feed-Forward Activation and Output sub-layers). To identify the main contributions of sub-layers to contextualisation, we first extract the sub-layer representations of polysemous words in minimally different sentence pairs, and compare how these representations change through the forward pass of the PLM network. Second, by probing on a sense identification classification task, we try to empirically localize the strength of contextualization information encoded in these sub-layer representations. With these probing experiments, we also try to gain a better understanding of the influence of context length and context richness on the degree of contextualization. Our main conclusion is cautionary: BERT demonstrates a high degree of contextualization in the top sub-layers if the word in question is in a specific position in the sentence with a shorter context window, but this does not systematically generalize across different word positions and context sizes. 



\end{abstract}

\section{Introduction}

\begin{figure*}[t]
\centering
\includegraphics[width=\linewidth]{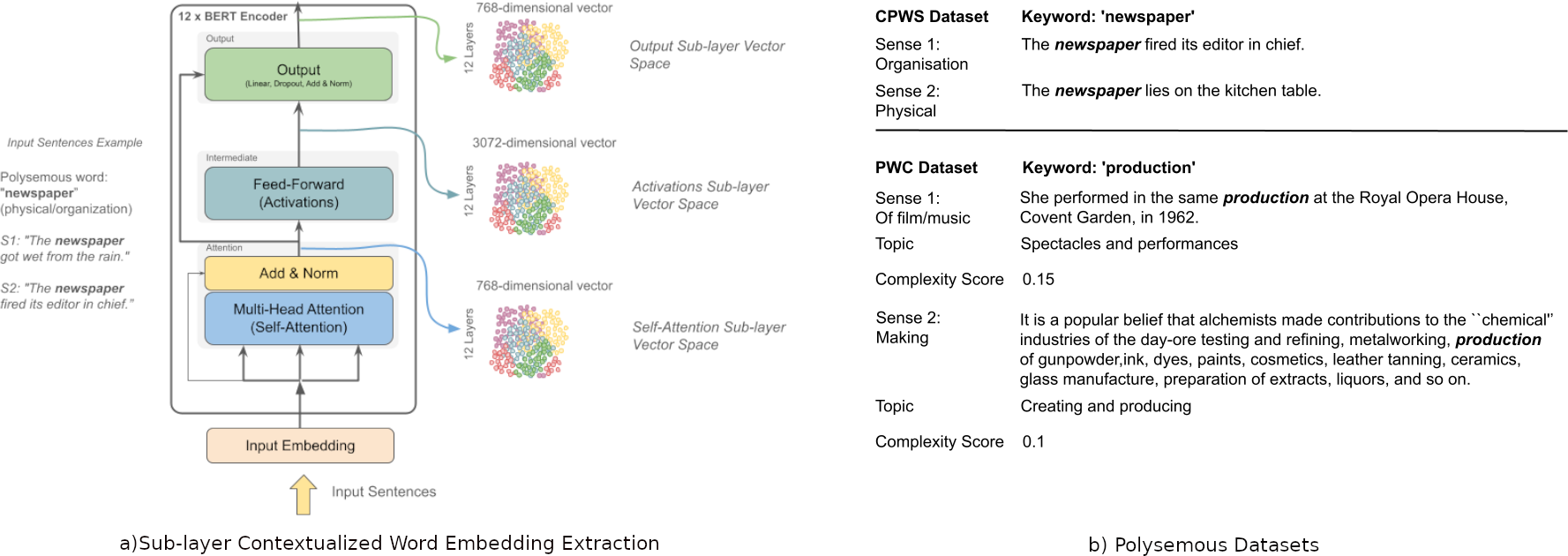}
\caption{\textbf{a) Extraction of contextualized word sub-layer latent representations from a BERT encoder layer:} From each BERT encoder layer, the Self-Attention (SA), Feed-Forward Activation (Acts) and Output sub-layer contextualized representations are extracted. \textbf{b)} Example Sentences in the CPWS - Contextualised Polysemy Word Sense v2 Dataset and PWC - Polysemous Word Complexity Dataset.}
\label{bert}
\end{figure*}

Contextualized representations of language  are a central element of modern era Natural Language Processing (NLP) Large Language Models (LLMs). Their pivotal role for PLMs has motivated researchers to quantify the amount of linguistic information encoded in such representations. Often, this quantification is achieved through empirical evidence using linear probing methodologies \cite{immer-etal-2022-probing,arps-etal-2022-probing}, where a linear supervised model is trained on such representations to predict the linguistic phenomenon of interest. As we are interested in contextualization phenomena, we look at the task of word sense disambiguation in polysemous words. Context plays a vital role in identifying the various senses of polysemous words. One of the most impactful encoder-only architectures that has been at the center of most studies for a long time and has successfully been shown to create contextualized word representations is the BERT model \cite{DBLP:journals/corr/abs-1810-04805,DBLP:journals/corr/abs-1909-00512,202403.0316}. Each encoder layer of BERT consists of three sub-layers producing latent representations: Self-Attention (SA), Feed-Forward Activation (Acts) and Output (Out) sub-layers \cite{DBLP:journals/corr/abs-1810-04805} (Figure \ref{bert}). These context-sensitive word representations are commonly known as Contextualized Word Embeddings (CWEs). We investigate the encoding of polysemy in these contextual representations by conducting fine-grained analysis on sub-layer latent representations in BERT. To investigate the degree of contextualization in PLM sub-layers, we use various similarity metrics (see Section \ref{met}). Our fine-grained sub-layer based investigation allows us to localize the degree of contextualization in the sub-layers of BERT.

Word Sense Disambiguation (WSD) is a long studied task in NLP and requires the identification of different senses of polysemous words. By using linear classifiers as sense probes for word sense identification of polysemous words, we empirically investigate the influence of word position and context length in the BERT sub-layer latent representations across BERT layers: As shown in Figure \ref{bert}b, the sentences in the upper part contain the polysemous word in a fixed position in the sentence, with only a minimal context conveying the sense of the word. In the lower part of the Figure, polysemous words appear in different positions of the sentence and are embedded in much longer contexts.  We measure the performance of each probe, which serves as empirical evidence for the contextualization within each sub-layer latent representation. Intuitively, the higher the performance (accuracy) of the linear classifier in a particular PLM sub-layer given the sub-layer latent representations, the better the sub-layer encodes word sense, thereby displaying a higher degree of contextualization of the polysemous word.


Our main contribution, unlike much previous work \cite{clark-etal-2019-bert,10.1162/tacl_a_00254,10.1145/3397271.3401075,zhao2020quantifying,ravfogel-etal-2020-unsupervised,10.1007/978-981-99-3878-0_36,10.1145/3397271.3401075,202403.0316}, is that we do not restrict ourselves to the output layer(s) of such networks, but we also investigate all different \textit{sub-layers} within each BERT encoder layer.  Our main findings from the experiments indicate the influence of context size on the degree of contextualization in BERT sub-layer representation. Our contributions are summarized as follows: 
\begin{itemize}
    \item We provide an in-depth finer-grained exploration of contextualization localization by conducting similarity-based investigations of how word representations change across layers and sub-layers of the PLM.
    


    \item We investigate the influence of word position/context-window settings on the degree of contextualization information encoded in the PLM sub-layers.
\end{itemize}



\section{Related Work}

There exists much research on using linear probes to quantify diverse linguistic knowledge encoded in neural language models
\cite{lin-etal-2019-open,merlo-2019-probing,lepori-mccoy-2020-picking,immer-etal-2022-probing,arps-etal-2022-probing}. \citet{ravichander-etal-2020-systematicity} examines hypernymy knowledge encoded in BERT representations using a consistency probe and observe that success of this probe on a hypernymy probing benchmark does not correspond to a systematic conceptual understanding of the underlying phenomena in BERT. \citet{DBLP:journals/corr/abs-2004-12198} find that the strongest contextualization interpretation effects occur in lower layers whereas the top layer do not contribute much to the contextualization. They also study the length of the context window that BERT layers effectively integrates for interpreting a word (a 10-word context window). In contrast, our focus is on probing the \textit{sub-layer representations} of BERT and understanding the effect of word position/context window in these sub-layers for polysemous words. 



For Word Sense Disambiguation, there exists various benchmark datasets and research that are part of various shared tasks \cite{raganato-etal-2017-word}. \citet{yenicelik-etal-2020-bert} investigate polysemy organization through separability and clusterability, which is similar to our work, but with a focus only on the last output layer of BERT. They observe that the subspace organization is determined by the intertwining of linguistic concepts such as semantics, syntax and sentiment and create closed semantic regions that seamlessly transition from one to another. 


\begin{figure*}[t]
\includegraphics[width=\linewidth]{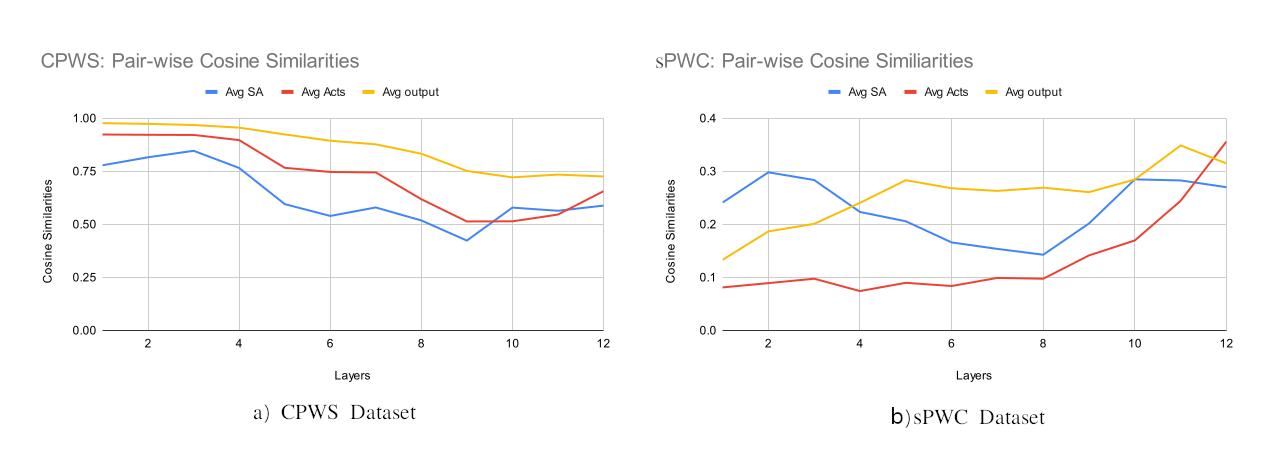}
\caption{\textbf{Pair-wise Polysemous Word Average Cosine Similarity} a) CPWS and, b) sPWC Dataset pair-wise average cosine similarity for Self-Attention (SA), Activation (Acts) and Output (output) sub-layers.}  \label{fig2}
\end{figure*}

\section{Methodology}
\label{meth}

\begin{figure*}[t]
\includegraphics[width=\linewidth]{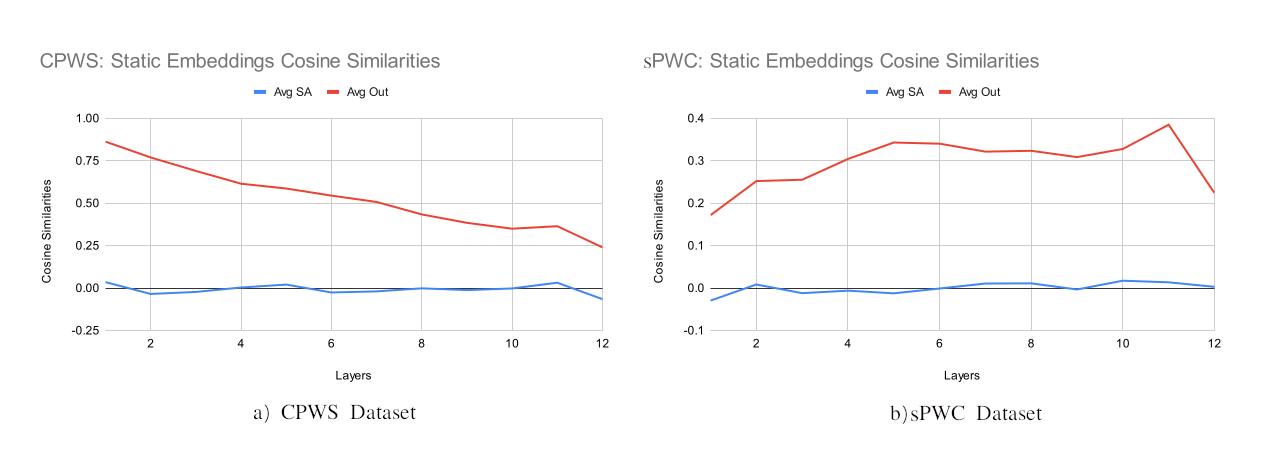}
\caption{\textbf{Static-Embeddings Average Cosine Similarity} a) CPWS Dataset and, b) sPWC Dataset static embeddings average cosine similarity for Self-Attention (SA), Activation (Acts) and Output (output) sub-layers.}  
\label{fig3}
\end{figure*}

\begin{figure*}[t]
\includegraphics[width=\linewidth]{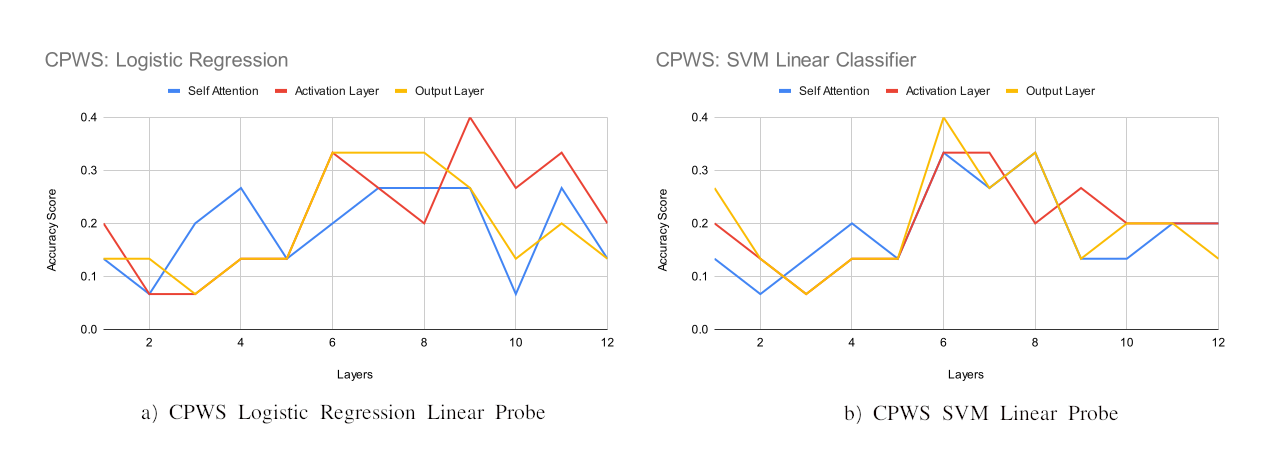}
\caption{\textbf{Linear Sense Probes: Logistic Regression (LR) and Support Vector Machine (SVM) Linear Classification Accuracies:} a) LR and b) SVM  BERT layer-wise linear sense probe accuracies on CPWS Dataset for Self-Attention (SA), Activation (Acts) and Output (output) sub-layers.}  
\label{fig4}
\end{figure*}

We conduct two steps to pinpoint the strength of localization across layers and sub-layers and to empirically investigate its influence on a word sense identification classification task.
\begin{enumerate}
    \item We first compare the representations of pairs of polysemous words with two different meanings in two different contexts across sub-layers of a pre-trained language model. This sheds light on the degree of contextualization that the model exhibits in sub-layers. Intuitively, while the word representations are equal at the very beginning (after the embedding layer), they become more dissimilar as they are altered by the transformer.
    \item We then probe the different sub-layer representations on a word sense disambiguation task to empirically investigate where the strongest contextualization is exhibited, as exemplified by a higher probe performance. Here, we test on different datasets with varying word positions and context sizes (long sentences vs very short ones), to also shed light on the influence of such data settings.
\end{enumerate}

\subsection*{Sense Probing} 
We formulate word sense identification of polysemous words as a classification task in a multi-label setting. Given the pre-trained BERT encoder, we take the CWEs produced by the sub-layers of each pre-trained encoder layer as the sense representation. Our dataset consists of the multiple senses for each polysemous word. Hence, we define a multi-class linear probing task with the senses as labels and the extracted sub-layer CWE representation of the respective polysemous word. We employ the One-vs-Rest strategy where the multi-class classification is split into one binary classification problem per class. We use two classifiers as our sense probes: A Logistic Regression (LR) linear classifier and a Support Vector Machine (SVM) linear classifier. 

By probing for senses of the polysemous words in their contextual sub-layer representations, we determine which sub-layers use this context to determine the senses. We define different sense probes for the three datasets: CPWS LR, CPWS SVM, sPWC LR, sPWC SVM, PWC LR and PWC SVM (see Section \ref{datasets} and \ref{results}). We do this for each sub-layer in a 12 layer BERT. This results in thirty six linear probes for each dataset \cite{DBLP:journals/corr/abs-1810-04805}. Each sense probe is trained and tested with the standard 80-20 train-test split. The accuracy score indicates the performance of detecting the senses in each sub-layer and across the 12 BERT encoder layers.

\section{Experimental Setup}

\subsection{Datasets}
\label{datasets}
We use the Contextualised Polysemy Word Sense v2 (CPWS) Dataset which contains custom samples of polysemous words in sentential contexts \cite{haber-poesio-2020-word}. This dataset contains sentences with a standard structure for each polysemous word, i.e, each polysemous word is in the second position after the definite determiner `The’. These sentences are characterized by short 
and natural contexts that invoke a certain sense of the polysemous word with a right context window. For example, the sentence "The newspaper fired its editor in chief" has the polysemous keyword "newspaper" (company, a copy of the paper, etc.) in the second position (see Figure \ref{bert}b for more examples).  

For investigating the influence of position of the keyword or context window, we combine two datasets: the Complex Word Identification (CWI) dataset \cite{yimam-etal-2017-cwig3g2} and the Sense Complexity Dataset (SeCoDa) \cite{strohmaier-etal-2020-secoda}. CWI consists of mixture of a professionally and non-professionally written news (WikiNews) and Wikipedia articles in English. This dataset consists of 34,879 samples. The SeCoDa dataset consists of the CWI dataset re-annotated with word senses. The dataset contains 1432 unique tokens with each token consisting of multiple senses. 
Since we are interested in polysemous words, we extract the tokens which have multiple senses along with their sense, context, and topics. We form a combined dataset by appending the polysemous words from the CWI dataset with its respective word senses in the SeCoDa dataset. We call this the Polysemous Word Complexity (PWC) dataset (see Figure \ref{bert}). The sentences in this dataset have longer context than the sentences in the CPWS dataset. 

The PWC dataset has label imbalance for different senses for each polysemous word. This can impact the linear probe performance trained for our polysemous sense prediction task (see Section \ref{meth}). To investigate the impact of this imbalance, we create a subset of our PWC dataset by extracting only one sentence for each sense, and we refer to this dataset as subset-Polysemous Word Complexity (sPWC). The similarity measures are used only in the CPWS and sPWC dataset as they contain pairs of senses for each polysemous word.

\subsection{The BERT Model}
We use a 12 layer BERT-base-uncased model as a representative language model. Each BERT layer consists of three sub-layers: Self-Attention (SA) sub-layer, Feed-Forward Activation (Acts) sub-layer, and Output sub-layer (see Figure \ref{bert}). 

\paragraph{\textbf{Self-Attention sub-layer:}} The Self-Attention sub-layer is a mapping of a query and a set of key-value pairs to an output, as computed below \cite{DBLP:journals/corr/abs-1810-04805,ferrando2024primerinnerworkingstransformerbased}:

\begin{equation}
SA(Q,K,V) = softmax(\frac{QK^T}{\sqrt{d_k}})V
\end{equation}
where, \textit{SA} is Self-Attention, \textit{Q,K,V} are query, key and value matrices, respectively. Large values of $d_k$ cause the dot product to grow large in magnitude, moving the $softmax$ to regions with extremely small gradients. To counteract this effect this dot product is scaled by $\frac{1}{\sqrt{d_k}}$. 

The BERT model consists of multi-head attention, which allows the model to jointly attend to information at different positions.

\paragraph{\textbf{Feed-Forward Activation sub-layer:}} The fully connected point-wise Feed-Forward activation sub-layer consists of two linear transformations with a ReLU activation in between \cite{DBLP:journals/corr/abs-1810-04805}:

\begin{equation}
FFN(x) = max(0,xW^l_{in} + b_1)W^l_{out} + b_2
\end{equation}
where, $W^l_{in}$, $W^l_{out}$ are two learnable input and output weight matrices at layer \textit{l} and $b_1$, $b_2$ are the respective biases.

\paragraph{\textbf{Output sub-layer:}} The Output sub-layer in each encoder consists of a linear layer, a dropout layer and a normalization layer.

We feed two sentences (with different senses) for each polysemous word to BERT and extract sub-layer vector representations for each BERT layer. For each word, we arrive at a set of vectors: \textit{Self-Attention (SA) sub-layer, Activation (Acts) sub-layer, and Output sub-layer}. We also extract the BERT static word embeddings (layer-0) for measuring how contextualised sub-layer representations diverge.

\subsection{Metrics}
\label{met}

\paragraph{\textbf{Sub-Layer Similarity:}} Let \textit{w} be the polysemous word that appears in a pair of sentences $\{\textit{s1, s}2\}$ at position \textit{i,j} in its respective sentence, $\{x_{s1}$, $x_{s2}\}$ be the sub-layer vector representations of the model \textit{m}. The sub-layer similarity of word \textit{w} in layer \textit{l} is:

\begin{equation}
SubLayerSim_{x}(w^l_{i,j}) = \frac{cos(x^l_{i_{s1}}, x^l_{j_{s2}})}{||x^l_{i_{s1}}|| ||x^l_{j_{s2}}||}
\end{equation}
where \textit{x} is Self-Attention, Feed-Forward Activation or Output sub-layer and $l = \{0,1,...,11\}$. 

\paragraph{\textbf{Static Word Embedding Similarity:}} For each word, we determine the cosine similarity between each sub-layer and its respective static word embedding from layer 0 (denoted as \textit{WESim}).

The \textit{Sub-Layer Similarity} and \textit{Static Word Embedding Similarity} for the respective sub-layer representations capture how the representations change through the forward pass of the PLM sub-layers. For example, for a given polysemous word \textit{w}, 
a low value of \textit{Sub-Layer Similarity} indicates higher degree of contextualization in the respective sub-layer vector representations.

\paragraph{\textbf{Principal Components Analysis (PCA):}} For qualitative analysis of the high-dimensional sub-layers (12 x 768 for Self-Attention and Output sub-layers, 12 x 3072 for Feed-Forward Activation sub-layers), we reduce them into two principal components using the PCA technique. PCA preserves the actual relative distance in the Euclidean data space and Principal Components (PCs) capture the direction of maximum variance. We determine squared L2 distances of the PCs between reduced sub-layer vector representations in the pair of sentences. We use these distance measures to quantitatively confirm the observations made using \textit{SubLayerSim} similarities.






\begin{figure*}[t]
\centering
\includegraphics[width=\linewidth]{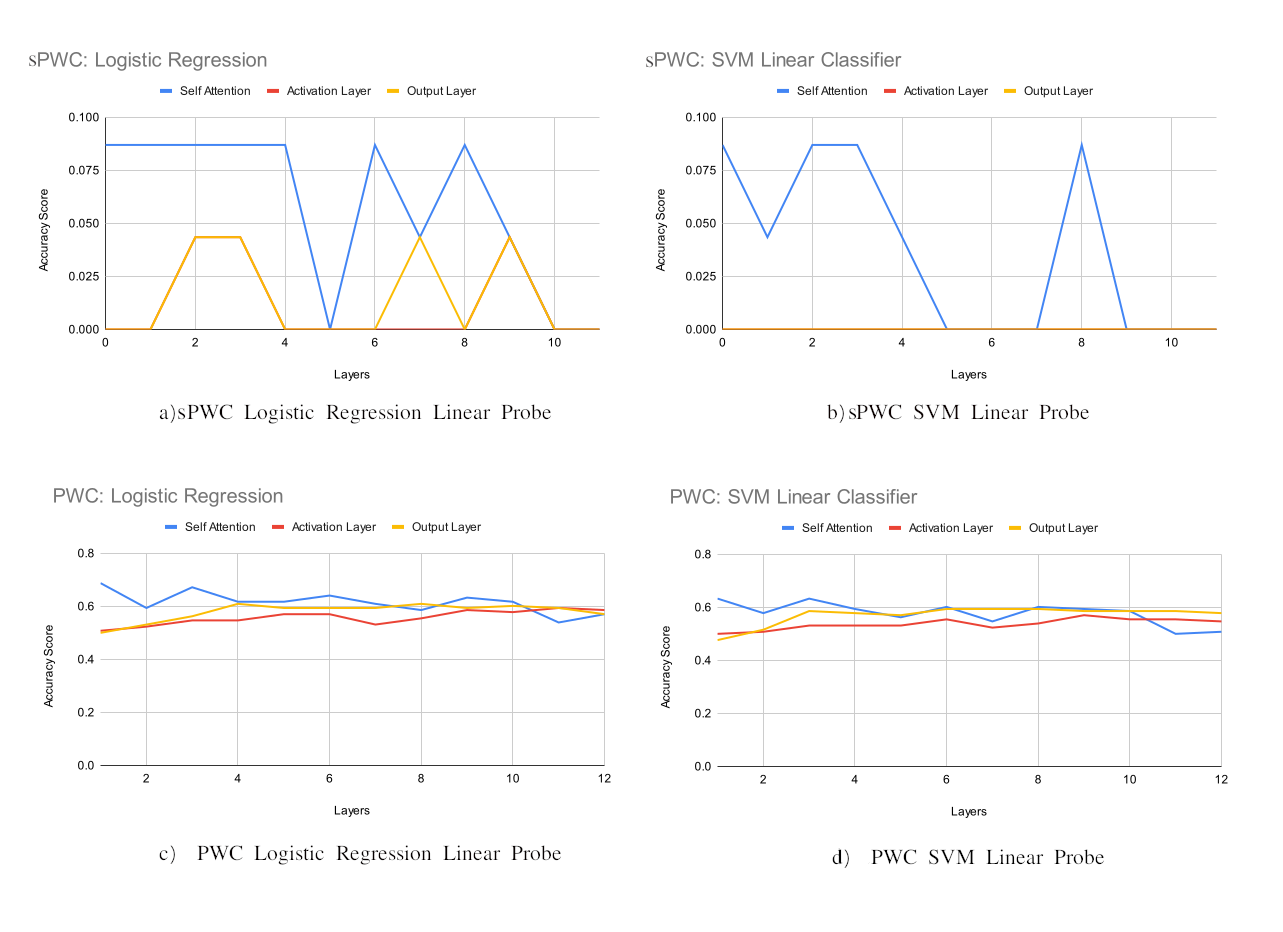}
\caption{\textbf{Linear Sense Probes: Logistic Regression (LR) and Support Vector Machine (SVM) Linear Classification Accuracies:} a,b) sPWC Dataset and c,d) PWC Dataset BERT layer-wise linear sense probe accuracies for Self-Attention (SA), Activation (Acts) and Output (output) sub-layers.}  
\label{fig5}
\end{figure*}

\section{Results \& Discussion}
\label{results}

\subsection{Contextualization}
\paragraph{Sub-layer Similarities:} We examine the similarities between polysemous words in different contexts. 
Figure \ref{fig2} presents the averaged cosine similarities for pair-wise sub-layer cosine similarities for each polysemous word. We observe that for shorter context windows (Figure 2a), the Output sub-layer similarity is closer to one, indicating a lower degree of contextualization as compared to Acts and SA sub-layers. The Acts and SA sub-layers are closer to each other, with higher contextualization in the SA sub-layer. On the other hand, for the shorter context window (Figure \ref{fig2}a), the sub-layers exhibit a higher degree of contextualization in the upper BERT layers whereas for longer context window (Figure \ref{fig2}b), each sub-layer behaves differently. For longer context windows in Figure \ref{fig2}b, we observe that the activation layer shows the highest dissimilarity in the lower BERT layers compared to self-attention and output sub-layers. 

The most interesting observation is in the static embedding cosine similarities, i.e. the cosine similarity between the static word embedding of the polysemous word (layer 0) and all other layers of BERT, presented in Figure \ref{fig3} for CPWS and sPWC. A notable finding is that SA sub-layer static embedding cosine similarities remain relatively consistent across BERT Layers. For output sub-layers, we observe that the degree of contextualization shifts from lower BERT layers to upper BERT layers as the context-window varies. In both datasets, the output sub-layers have a higher similarity than its respective SA sub-layer. This could indicate the influence of residual connections in the output sub-layer \cite{vijayakumar2023exactly}. 
The major difference between both datasets is the position of the keyword and length of the context windows (see Figure \ref{bert}b). \citet{DBLP:journals/corr/abs-1909-00512} observes that CWEs are dissimilar to each other in upper BERT layers. Our empirical analysis shows that this observation holds only for the polysemous words that are part of the shorter contexts, whereas with the longer contexts, polysemous words do not exhibit such behaviour in the lower BERT layers. Similar to findings in ELMo embeddings using canonical co-prediction examples, \citet{haber-poesio-2020-word} observe that the target word position and the function significantly impact sense shifting, potentially overshadowing other factors. We assume that domain and text complexity may also influence this behavior. This observation leads us to a cautionary conclusion that the keyword-position/context-windows impacts the degree of contextualization and that the degree of contextualization in lower and upper BERT layers cannot be generalized across keyword positions and context lengths.


\begin{table}
\centering
\begin{tabular}{llll}
\hline
 & \textbf{Avg Sa} & \textbf{Avg Acts} & \textbf{Avg Outs} \\
\hline
\textbf{SLSim} & 0.6329 & 0.7309 & 0.8614\\
\textbf{WESim} & 0.008 & - & 0.521 \\
\textbf{L2 Dists} & 3.217 & 3.413 & 1.195\\
\hline
\end{tabular}
\caption{SLSim: SubLayerSim, WESim: Word Embeddings and L2 distances for each sub-layer and all words.}
\label{tab:allwordsavg}
\end{table}

\subsection{Sense Probing}
Intuitively, if the sense classifier succeeds, it means that the pre-trained encoder sub-layer contains sense information and higher accuracy score means that the particular BERT encoder sub-layer encodes the senses more accurately relative to the other sub-layers. 

The accuracies of the probing task using LR and SVM on all the three datasets are shown in Figure \ref{fig4} and \ref{fig5}. For the CPWS dataset, both linear sense probe accuracies indicate that most information regarding the different senses of the polysemous words is encoded in middle and upper BERT layers (Figure \ref{fig4}a and b). This observation is consistent with the presence of a higher degree of contextualization observed earlier (Figure \ref{fig2}a). Interestingly, the performance of the probes on sPWC dataset (singe sentence per word and sense) are very poor and is much harder to extract any meaningful reasons and conclusions (Figure \ref{fig5}a and b). A known behavior of probing tasks is that probes fail to adequately reflect differences in representation on large training data and require a reduced amount of probe training data to show different accuracies with respect to pre-trained representations \cite{voita-titov-2020-information}. Despite the relatively small dataset size of our sPWC, the linear sense probe fails to perform. Finally, in Figure \ref{fig5}c and d,  both linear sense probes perform consistently with similar accuracies on all encoder sub-layer representations. We suggest that the accuracies do not vary and do not reflect the various sense encodings due to label imbalance and the larger size of the PWC dataset. 

\begin{figure*}[t]
\centering
\includegraphics[width=\linewidth]{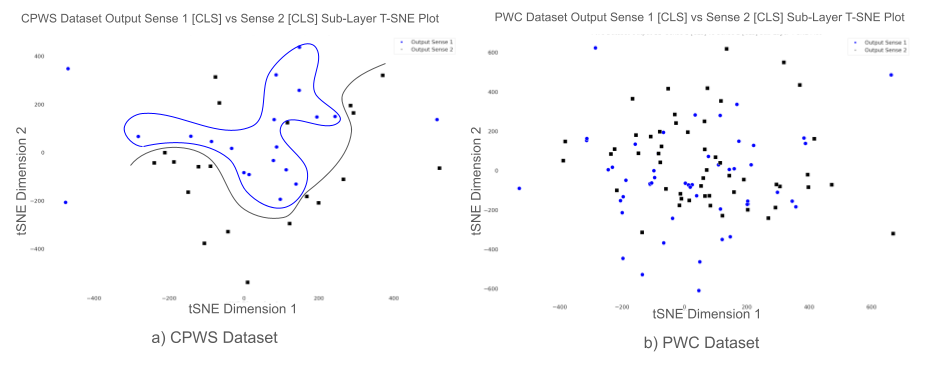}
\caption{\textbf{CPWS vs PWC Dataset T-SNE Output Sub-Layer [CLS] Sense 1 vs Sense 2: } T-SNE plots of two senses of CPWS and PWC Dataset polysemsous words for Output sub-layer CLS token from last BERT layer (layer 12).}  
\label{out-cls-tsne}
\end{figure*}

Another reason for this low performance could be that BERT is originally pre-trained on the BooksCorpus \cite{DBLP:journals/corr/ZhuKZSUTF15} while our dataset consists of news articles. This domain difference may to some extent cause the representations to not sufficiently encode the different senses of polysemous words. \citet{pimentel-etal-2020-pareto} explain the trade off between accuracy and complexity of linear probes and we assume that the linear probes are not able to capture the encoded sense in larger context-windows. Another interesting observation is that CWEs are anisotropic, that is they are not uniformly distributed with respect to direction and they do not correspond to a finite-number of word-sense representations \cite{DBLP:journals/corr/abs-1909-00512}. We suggest this could also be a reason for the linear sense probes to perform poorly on the longer context-window dataset.

\subsection{Qualitative Analysis}

\paragraph{PCA:} We observe that the PCA bi-plots for SA, Acts and Output sub-layers are structurally different, indicating different structural alignment in their respective high-dimensional spaces. Examining the per-word average L2 distances, we observe that the Output sub-layer L2 distances are much lower than the respective SA and Acts sub-layers, indicating a stronger contextualization in the SA and Acts sub-layers (Table \ref{tab:allwordsavg}). 

We additionally examine the CLS token representation for the last BERT layer using T-SNE plots \cite{JMLR:v9:vandermaaten08a}. We observe that the different senses form separate clusters for CPWS dataset while they do not do this for the PWC dataset (see Figure \ref{out-cls-tsne}). Similar observations are made by \citet{wiedemann2019does}, where BERT embedding space shows some senses form clearly separable clusters.

\section{Conclusion}

In this paper, we present a methodology for in-depth finer-grained empirical analysis of Contextualized Word Representations (CWEs) in transformer-based masked language models. We hold the model architecture constant and investigate the impact of word position/context windows on the word sense identification in sub-layers. Using one of the most common interpretability methods, linear probing, we investigate the degree of contextualization these language model sub-layers encode and localize this contextualization to the respective encoder sub-layers, by conducting empirical studies. 

We draw limited and cautionary conclusions on the degree of contextualization that BERT sub-layers encode for polysemous words. The context plays a vital role in determining the different polysemous word senses. Interestingly, we observe varying trends in lower and upper BERT encoder sub-layers when input polysemous keywords have different position/context-windows. As in much previous research, we employ the most common interpretability method, linear probing, for conducting empirical experiments.
Generally, in previous works, the goal of a probing task is to test if contextual representations encode certain linguistic properties and the contextual representations are extracted from the last layer or the output layer of the encoder \cite{DBLP:journals/corr/abs-1909-00512,zhao2024opening}. In our study we extend our investigations to a fine-grained analysis by extracting contextualised encodings from three sub-layers: \textit{Self-Attention sub-layers, Feed-Forward Activation sub-layers and Output sub-layers} for all the 12 encoder BERT layers.   

Based on our experiments, we find evidence suggesting the following trends. Shorter context windows (limited on the left window) lead to higher contextualization in upper BERT layers for all sub-layers whereas longer context windows show different behaviour in each BERT sub-layer. This is supported by the performance of linear sense probing tasks on the CPWS dataset. The performance of these linear sense probes on the dataset with longer context windows (sPWC) and the larger dataset (PWC) does not show conclusive evidence of the word sense encoded in the respective sub-layers. We suggest this observation could be due to the sub-optimal performance of linear probes and the extracted contextual sub-layer representations not entirely capturing the different word senses.
Using the sPWC/PWC dataset, in our future research we study the impact of fine-tuning and context augmentation in these sub-layers.

\section*{Limitations}
We highlight a few limitations of our experiment settings and methodology. Firstly, in our experiments, we focus on the impact of word position and its context window in differentiating the respective senses. We have not considered the syntactic structure, like subject verb agreement, of the sample sentences. Second, we use the sub-layer representations of BERT-base-uncased 12 layer model as our representative constant model architecture. The sense probes presented are model agnostic, nevertheless, we are yet to conduct generalizability studies on other models. Third, we use very simple linear probes and we are yet to explore if complex probes performs better than simple linear probes. Recent methods aims at balancing probe performance with probe complexity \cite{hewitt-liang-2019-designing,pimentel-etal-2020-pareto}.



\bibliography{acl_latex}

\appendix

\newpage
\section{Dataset Statistics}
\label{sec:appdatasets}

Table \ref{tab:datstat} shows the statistics of our datasets. The \textit{Unique Keywords} column indicate the total number of unique polysemous words that is present in the respective dataset.

\begin{table}
\centering
\begin{tabular}{lll}
\hline
 \textbf{Dataset} & \textbf{Total Samples} & \textbf{Unique Keywords} \\
\hline
\textbf{CPWS} & 58 & 29 \\
\textbf{sPWC} & 228 & 114  \\
\textbf{PWC} & 34,879 & 1432 \\
\hline
\end{tabular}
\caption{CPWS: Contextualised Polysemy Word Sense v2, sPWC: subset-Polysemous Word Complexity, PWC: Polysemous Word Complexity dataset statistics.}
\label{tab:datstat}
\end{table}

\end{document}